\documentclass[conference]{IEEEtran}
\IEEEoverridecommandlockouts
\usepackage{amsmath,amssymb,amsfonts}
\usepackage{subcaption}
\usepackage{algorithmic}
\usepackage{graphicx}
\usepackage{textcomp}
\usepackage{xcolor}
\def\BibTeX{{\rm B\kern-.05em{\sc i\kern-.025em b}\kern-.08em
    T\kern-.1667em\lower.7ex\hbox{E}\kern-.125emX}}

\usepackage[hidelinks]{hyperref}
\usepackage[sorting=none]{biblatex}
\begin{document}

\title{Terrain Classification using Transfer Learning on Hyperspectral Images: A Comparative study\\
}

\author{ Uphar Singh, Kumar Saurabh, Neelaksh Trehan, Ranjana Vyas, O.P. Vyas \\Indian Institute of Information Technology, Allahabad, India \\ pse2017003@iiita.ac.in, pwc2017001@iiita.ac.in, iim2017002@iiit.ac.in, ranjana@iiita.ac.in, opvyas@iiita.ac.in
}

\maketitle

\begin{abstract}
A Hyperspectral image contains much more number of channels as compared to a RGB image, hence containing more information about entities within the image. The convolutional neural network (CNN) and the Multi-Layer Perceptron (MLP) have been proven to be an effective method of image classification. However, they suffer from the issues of long training time and requirement of large amounts of the labeled data, to achieve the expected outcome. These issues become more complex while dealing with hyperspectral images. To decrease the training time and reduce the dependence on large labeled dataset, we propose using the method of transfer learning. The hyperspectral dataset is preprocessed to a lower dimension using PCA, then deep learning models are applied to it for the purpose of classification. The features learned by this model are then used by the transfer learning model to solve a new classification problem on an unseen dataset.  A detailed comparison of CNN and multiple MLP architectural models is performed, to determine an optimum architecture that suits best the objective. The results show that the scaling of layers not always leads to increase in accuracy but often leads to overfitting, and also an increase in the training time.The training time is reduced to greater extent by applying the transfer learning approach rather than just approaching the problem by directly training a new model on large datasets, without much affecting the accuracy.
\end{abstract}


\section{Introduction}
Hyperspectral image (HSI) is an image cube made up of hundreds of spatial images. For collecting hyperspectral images, a hyperspectral camera is used to collect a data cube. A normal image has 3 channels of pixel values namely, Red, Green and Blue (RGB). While in the case of hyperspectral images, tens to hundreds of narrow colour channels (i.e. wavelengths) are recorded. By combining all these narrow colour channels we are able to process different qualities of our target image. This forms our basis of classification of Hyperspectral images. The high dimensionality and lack of data samples make these images a highly complex task to classify.\vspace{0.3cm}\\
In their survey, \textit{Shao et al.}, have very well explained the application of transfer learning in the context of visual categorization \cite{tlSurvey}. The training data in the same feature space or with same distribution as the future data can’t surely avoid the over-fitting, when the availability of labeled data is low. Transfer learning can help in such situations as related data in different domains can be used to expand the knowledge. Transfer learning helps by extracting useful information from data in a related domain and  transferring  it to be used in the target  domain.\vspace{0.3cm}\\
The convolutional neural network (CNN) is a class of deep learning neural networks. CNNs represent a huge breakthrough in image recognition. The  use  of CNN for HSI classification is also visible in recent works \cite{3d2dCNN}. They’re most commonly used to analyze visual imagery and are frequently working behind the scenes in image classification. Making use of CNN would be a brilliant way to implement the classification.\vspace{0.3cm}\\
While CNNs have become a standard model for the image classification, the Multi-Layer Perceptron (MLP) models are also extensively used for the HSI classification, especially when the computing resources are limited \cite{mlpIntro}. MLP classification is recommended for the embedded platforms, as it showcased the shorter run-time as compared to CNN and comparable accuracy.\vspace{0.3cm}\\
However, in the case of HSIs, training time is long for the deep learning models, and the demand for the marked data is large. To overcome this issue, we try transfer learning for the problem. Deep learning models are trained on some other dataset and using the learned parameters, the image in the given dataset is classified. This would reduce the training time as well as help us overcome the obstruction of having fairly small dataset given the high dimensionality.
With the advent of modern remote sensing technology it has become possible to obtain much higher resolution and much better quality satellite images than ever before, as such it is important to scale the software architecture that processes these images to achieve maximal utilization. One category of satellite images is hyperspectral images, these work similarly to ordinary images in that they encode the information available as multiple values which correlate to specific channels of specific pixels of the image, i.e. an image consists of pixels which consist of a set of multiple numeric values. One of the primary differences between hyperspectral images and ordinary images is that ordinary images consist of only three channels: red, green and blue, whereas hyperspectral images consist of many more channels.\\
Ordinary images are like this due to the fact that human eyes have limits to their capabilities of perceiving wavelengths of the electromagnetic spectrum. However, digital processing tools are not bound by the same limitations, and hence can process the information contained within hyperspectral images and produce greater knowledge about entities within the image. All this intrigues to further explore the hyperspectral images. 

Hyperspectral Image is an image captured in such a way that each pixel includes complete spectrum. Thus the Hyperspectral Image forms a three dimensional data (i.e. is in the form of a data cube). In the Hyperspectral Dataset, we have an image along with its correct labels, also known as ground truth values.\\
Since the availability of the labeled Hyperspectral data is low, so to avoid overfitting and also to reduce the training time, we explore the transfer learning approach. Using related data from a different domain, we expand the knowledge for the problem domain.\vspace{0.3cm}\\
The following datasets are used:
\begin{itemize}
    \item \textbf{The Indian Pines (IP)} \cite{indianPines}:\\ This dataset has an image with 145x145 spatial dimensions and 224 spectral bands in the wavelength range of 400 to 2500 nm, out of which 24 spectral bands covering the region of water absorption have been discarded. The ground truth available is designated into 16 classes of vegetation.
    \item \textbf{The University of Pavia (UP)} \cite{paviaUniversity}:\\ This dataset consists of an image with 610x610 spatial dimension pixels with 103 spectral bands in the wavelength range of 430 to 860 nm. The ground truth is divided into 9 urban land-cover classes.
\end{itemize}
The deep learning models will be trained for the classification of the Indian Pines dataset. These models will then be used to transfer the knowledge extracted from the Indian Pines dataset, i.e. they will act as a pre-trained model for the classification task of the University of Pavia dataset.\vspace{0.3cm}\\
The project aims to compare the different deep learning models coupled with transfer learning, to perform classification of the Hyperspectral Images (the University of Pavia image in our case).

\section{Literature Review}
With the advent of remote sensing technology, the popularity of Hyperspectral image (HSI) classification has increased. Hyperspectral data possess complex characteristics which transform the task of accurate classification into an interesting and challenging problem especially when restricted to traditional machine learning methods.\\
Recent developments in the field of machine learning have identified deep learning to be a powerful tool to achieve feature extraction and address nonlinear problems, furthermore deep learning models are already heavily used in several image processing applications. Considering these factors, and especially considering the achievements of deep learning techniques in other, similar applications, deep learning has also been introduced to classify HSIs. Deep learning methods have demonstrated adequately impressive performance. In their survey paper, \textit{Ying Li et al.}, have systematically reviewed several pixel-wise and scene-wise Remote Sensing image classification approaches which use DL methods \cite{dlSurvey}. The authors performed comparative analysis on Convolutional neural networks, Stacked autoencoders and Deep belief networks. Their paper involved analysis of Spectral feature classification, Spatial feature classification and Spectral-spatial feature classification, their results on the Pavia University dataset show CNNs with Spectral-spatial feature classification to be most accurate. Additionally they found dual channel CNNs to provide the highest accuracy on Indian Pines dataset.\vspace{0.3cm}\\
Different deep learning models may be applied for the classification of hyperspectral images that capitalise on both, the spectral and the spatial signatures of the hyperspectral image to obtain state of the art accuracies. A detailed comparative study of the various deep learning models used for hyperspectral classification including deep belief networks(DBN), Generative Adversarial Networks(GAN’s), Recurrent Neural Networks(RNNs) and Convolutional Neural Networks(CNNs) and it’s different variants like the Gabor-CNN, S-CNN and RESNET, has been presented by \textit{Shutao Li et al.} \cite{dlHSI}.\\
It was observed that the CNN based models(S-CNN,Gabor-CNN)  were particularly good for the classification task, with most of them having state of the art accuracies ($>$99\%) and apart from the 3-D GAN model all of the deep learning models had a satisfactory accuracy. It was also shown that the deep learning models outperform the traditional machine learning models like SVM, EMP which were also used for the purpose of classifying hyperspectral images.\vspace{0.3cm}\\
The Spectral-spatial features are extracted from a target pixel and its neighbours. The convolution on these spectral-spatial features leads to a number of 1D feature maps. The feature maps are stacked into a 2D matrix and can be fed to the Convolutional neural net. This type of model can be reffered as a HSI-CNN model \cite{HSI-CNN}. For using classifiers like KNN and  SVMs one requires a more in-depth knowledge of Hyperspectral images, on the other hand Deep Learning techniques like neural nets allow us to extract complicated features from hidden layers without too much pre-processing of the data.\\
HSI-CNN is a good trade-off between number of training samples and complexity of the network, and overcomes over-fitting. XGBoost can be considered as a substitution of the softmax layer of HSI-CNN in order to prevent over-fitting.\vspace{0.3cm}\\
The Multi-Layer Perceptron (MLP) have also been extensively used for classification of the multispectral as well as hyperspectral images. They are relatively fast to train and produce comparable results to the CNNs. A better performance is observed when they are coupled with the PCA \cite{mlpLit1}.\\
A study by Beatriz et al. has focused on the time optimization for the classification of HSIs by multilayer perceptron. They proposed 5 different architectures for MLP and compared them with the state of the art SVM model. They also proposed a parallel programming scheme for training the MLP \cite{mlpLit2}.\vspace{0.3cm}\\
A variation can be using SVM in the last layer of the CNN model instead of the traditional softmax layer (SVM-CNN model). The resultant model would now require a lesser training time in contrast to that with the softmax layer, which happens to be more computationally expensive due to the softmax function. The model with the softmax layer has been showed to slightly outperform the SVM-CNN in terms of accuracy, but the SVM-CNN is slightly faster to train \cite{SVM-CNN}.\vspace{0.3cm}\\
The paper by \textit{Roy et al.} shows the advantages of using Hybrid Spectral Net (Hybrid SN), a 3D-2D CNN instead of using pure 2D or pure 3D CNN \cite{3d2dCNN}. The pure 2D CNN on its own is not able to extract features regarding the spectral dimensions while pure 3D CNN would become computationally expensive to use and it performs worse for classes having similar textures over many spectral dimensions. The 3D-CNN and 2D-CNN layers are assembled for the proposed model (Hybrid SN) in such a way that they utilise both the spectral as well as spatial feature maps to their full extent to achieve maximum possible accuracy. To reduce the number of spectral bands (more accurately remove the spectral density) PCA is applied over the HSI image.\\
The HSI data cube is divided into small overlapping 3D patches, their truth labels are decided by the label of the centered pixel. The 2D convolution is applied once before the flatten layer to extract abstract high level features. The research paper confirms the superiority of the proposed method by performing experiments over three benchmark datasets and comparing the results with state of the art methods. The proposed model is shown to be computationally more efficient than the pure 3D CNN model.\\
The paper by \textit{Xuefeng Jiang et al.} \cite{markovTL} emphasizes a new technique of feature extraction using transfer learning which significantly lowers the training time of the CNN used and reduces its dependency on large labelled datasets. This leverages the Markov property of the images to separate images with class tags and  train the CNN on random band samples selected from the datasets.\\
The paper by \textit{Ke Li et al.} \cite{boltTL} explores another transfer learning approach for hyperspectral image classification using deep belief networks wherein a limited Boltzmann machine Network is trained on the source domain data and its first few layers are extracted to be used for the target domain network. The target domain network is fine tuned further and used for classification of images in the target domain. The number of layers to be transferred are also varied and chosen for best accuracy.\vspace{0.3cm}\\
CNN-based frameworks can achieve high accuracy in an object classification task. But when coupled with transfer learning, they can give high performance in detection tasks as well, while reducing the required training time and computing resources \cite{snowDetect}.\vspace{0.3cm}\\
Taking inspiration from these papers we use a transfer learning approach coupled with CNN and MLP models, and perform a comparative analysis in the end.

\section{Proposed Methodology}
\subsection{Dataset Preparation}
Two datasets, namely the Indian Pines and the Pavia University Hyperspectral datasets,  were used, the Indian Pines for training the CNN model from which the CNN stump (the convolutional layers and the pooling layers barring the fully connected layers) was extracted to be trained on the Pavia  dataset. Both images have different distributions that thus makes the applicability of Transfer Learning perfectly suitable.\vspace{0.3cm}\\
The data used for initial model training was the Indian Pines Hyperspectral Data collected by the AVIRIS sensor over the Indian Pines test site in North Western Indiana. It comprises 145x145 pixels and 224 spectral reflectance bands in the wavelength range 0.4-2.5 $\mu$m. Two thirds of the Indian pines is agricultural landscape and the remaining one-third contains perennial vegetation. The ground truth is designated into 16 classes as described by Table \ref{tab:indianPines}.

\begin{table}[h] 
\centering
\begin{tabular}{|l|c|c|}
\hline
\bfseries S.No. & \bfseries Class & \bfseries Samples \\
\hline
1 & Alfalfa & 46\\
\hline
2 & Corn-notill & 1428\\
\hline
3 & Corn-mintill & 830\\
\hline
4 & Corn & 237\\
\hline
5 & Grass-pasture & 483\\
\hline
6 & Grass-trees & 730\\
\hline
7 & Grass-pasture-mowed & 28\\
\hline
8 & Hay-windrowed & 478\\
\hline
9 & Oats & 20\\
\hline
10 & Soybean-notill & 972\\
\hline
11 & Soybean-mintill & 2455\\
\hline
12 & Soybean-clean & 593\\
\hline
13 & Wheat & 205\\
\hline
14 & Woods & 1265\\
\hline
15 & Buildings-Grass-Trees-Drives & 386\\
\hline
16 & Stone-Steel-Towers & 93\\
\hline
\end{tabular}
\caption{Ground-truth classes for the Indian Pines scene and their respective samples number}
\label{tab:indianPines}
\end{table}

The Pavia University dataset was used to test the model built using transfer learning. It was acquired by the ROSIS sensor over Pavia in northern Italy. It contains 103 spectral bands with 610x610 pixels. The ground truth values for the pixels cover 9 different classes. Table \ref{tab:paviaUniversity} describes the ground truth classes for the Pavia University dataset and their respective sample numbers.

\begin{table}[h] 
\centering
\begin{tabular}{|l|c|c|}
\hline
\bfseries S.No. & \bfseries Class & \bfseries Samples \\
\hline
1 & Asphalt & 6631\\
\hline
2 & Meadows & 18649\\
\hline
3 & Gravel & 2099\\
\hline
4 & Trees & 3064\\
\hline
5 & Painted metal sheets & 1345\\
\hline
6 & Bare Soil & 5029\\
\hline
7 & Bitumen & 1330\\
\hline
8 & Self-Blocking Bricks & 3682\\
\hline
9 & Shadows & 947\\
\hline
\end{tabular}
\caption{Ground-truth classes for the Pavia University scene and their respective samples number}
\label{tab:paviaUniversity}
\end{table}

Before model training both data sets were subjected to PCA for dimensional reduction, which reduces the spectral dimension to 30 for the Indian Pines and the Pavia dataset. This is done so as to facilitate faster training time for the models and reduces the memory cost.
\\It should be noted that before feeding the dataset for training we need to prepare multiple distinct image cubes from the images, and the size of the image cubes behaves as a hyper-parameter. Increasing the size of the cube though may increase the accuracy but will result in a drastic increase in the requirement of RAM. For this purpose the image cubes are taken of size 5x5x30. Further we also need to flatten the given input image after reducing the dimension using PCA, since unlike a CNN model where we can directly feed an image, a MLP model requires a flat matrix of input parameters.

\subsection{Model Training}
All of the following models were made in 2 phases. In first phase, the models were trained on the Indian Pines dataset, and then in second phase, transfer learning was applied and models were trained on Pavia University dataset.

\subsubsection{CNN Model}
Firstly, a CNN model was trained on the Indian Pines dataset. This CNN uses both 3-D and 2-D convolution operations to leverage feature map extraction from both the spatial and the spectral dimensions, in case of just the 2-D convolution however, the spectral information would not have been captured well enough for accurate classification. The model is trained with the 5x5x30 data-cubes obtained from pre-processing the data which is shuffled and split into a train to test ratio of 70:30.\vspace{0.3cm}\\
The first 3 layers are convolution 3-D layers, which consist of 8,16 and 32 filters respectively. After this, a convolution 2-D layer is applied which consists of 64 filters and the filter size of 3 by 3. This is the 4\textsuperscript{th} convolution layer. The output from this convolution layer is then flattened (i.e. converted all the activations into 1-D vector) and then fed into the Artificial Neural Net type architecture that mainly consists of fully connected dense layers with ReLU activations.We have 2 dense layers combined with dropout regularization of 0.4. We have made use of Leaky ReLU activation in the 2 dense layer, and finally an output layer that uses a Softmax Activation function for class prediction.We have used Adam Optimizer for training the CNN.\vspace{0.3cm}\\
Once the above model is trained on the Indian Pines dataset the CNN stump of the model i.e. the convolutional and pooling layers are extracted and the later layers i.e. the fully connected layers and the softmax layers are stripped. To this convolutional stump, a new set of layers is connected with 3 fully connected layers with ReLU activation and a softmax layer of 9 units are appended. We also use a dropout of 0.4 for the fully connected layers. The layers excluding the fully connected layers and the softmax layers are frozen so that the weights for those layers are not modified and training only takes place for the fully connected layers and the softmax layer. The patches obtained from the Pavia dataset are then split into the training and the test set in a ratio of 40:60, and trained for 3 epochs. 
\subsubsection{MLP Model 1}
\begin{figure}[ht]
  \centering
  \frame{\includegraphics[scale = 0.7]{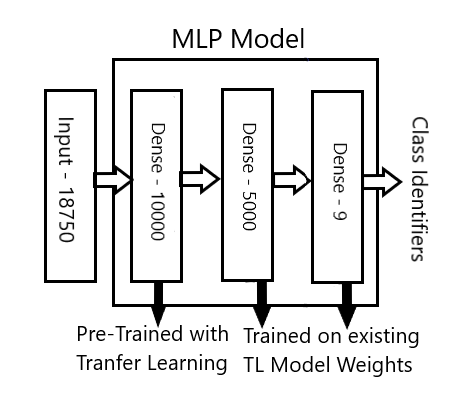}}
  \caption{Diagrammatic representation of MLP Model 1}
  \label{fig:mlpModel1D}
\end{figure}
The MLP model used here is a standard MLP model with three dense layers with 10000, 5000 and 9 perceptrons respectively and is represented diagrammatically in Fig \ref{fig:mlpModel1D}. The model firstly was trained on Indian Pines dataset and then the last two layers were deleted while making use of the model for prediction on Pavia dataset using transfer learning. The first and the second layers are activated using the ReLu activation function while the last layer that gives the Class identifiers is activated using a SoftMax activation function. The initial model was trained for 20 epochs with a batch size of 128, while on the other hand when the model was used for transfer learning and trained on Pavia dataset only 10 epochs with 128 batch size were performed. \\

We have used Adam Optimizer for training the MLP perceptron layers. Adam is an adaptive learning rate optimization algorithm that’s been designed specifically for training deep neural networks. Adam can be considered as a mixture of RMS prop and Stochastic Gradient Descent. It uses the squared gradients to scale the learning rate like RMS prop and it takes advantage of momentum by using moving average of the gradient instead of gradient itself like SGD with momentum. The Loss used is categorical cross entropy and the evaluation metrics is accuracy.

\subsubsection{MLP Model 2}
\begin{figure}[h]
  \centering
  \frame{\includegraphics[scale = 0.38]{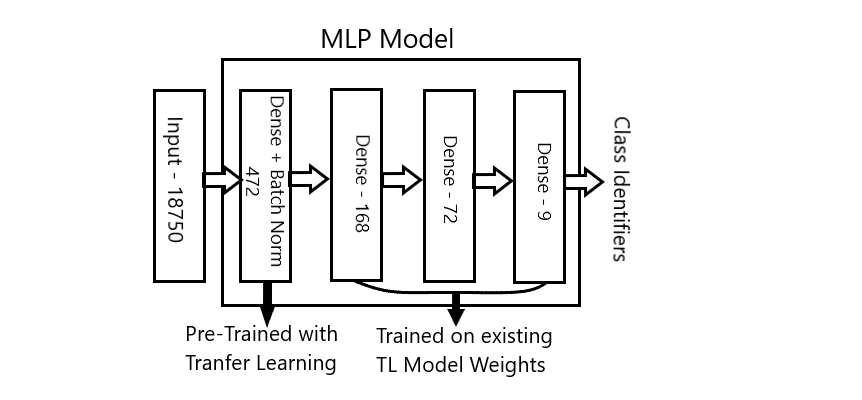}}
  \caption{Diagrammatic representation of MLP Model 2}
  \label{fig:mlpModel2D}
\end{figure}

Since there is no perfect model that can be defined for any Neural Network problem hence the feasibility of the solution and model depends on whether the error generated by the model is under the permissible parameters of our problem. During our literature survey we found that Terrain Classification on Hyperspectral Images has been under research at IIT Delhi with the association of DRDO. Though not much details about the methodology used was found for it being a closed project, but we used one of architecture of the MLP model that was used for their purpose. The model can be visualised diagrammatically as in Fig \ref{fig:mlpModel2D}.\\
In this model the initial training was done in the Indian Pines dataset with just two Hidden Layers. Later when Transfer learning was performed, the last layer along with its batch normalization extension was chopped off, and three more layers were added. The first layer is a Dense Layer of 472 perceptrons, followed by a Batch Normalisation layer. This first layer was left as it is during transfer learning, and three more layers were added with 168, 72 and 9 perceptrons respectively. These layers were the trainable layers that were trained during transfer learning. All these perceptron layers are activated using a ReLu activation function and the optimization technique used while training is Adam Optimization, since it is believed to outperform other optimization techniques like RMSProp. 

The initial model was trained for 20 epochs with a batch size of 128, while on the other hand when the model was used for transfer learning and trained on Pavia dataset only 10 epochs with 128 batch size were performed.

\subsubsection{MLP Model 3}
\begin{figure}[h]
  \centering
  \frame{\includegraphics[scale = 0.43]{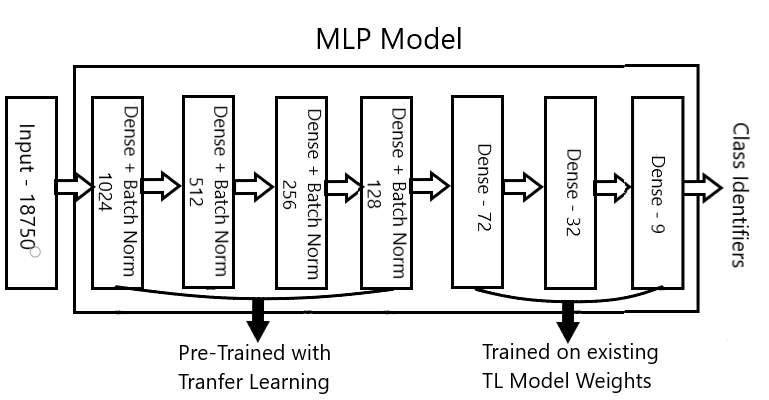}}
  \caption{Diagrammatic representation of MLP Model 3}
  \label{fig:mlpModel3D}
\end{figure}

This model is an extension of the Model 2 with a greater number of pre-trained layers but with lesser number of perceptrons, as models also need to be compared by scaling them horizontally, i.e., adding more layers but reducing the strength of each layer. The visual representation of the model is give in Fig \ref{fig:mlpModel3D}.\\
This model was also first trained with 5 Hidden layers on Indian Pines dataset, but when transfer learning was applied its last two layers were chopped off, and three new dense layers were added. The layers from the saved models were not trained on the Pavia dataset, while the three dense layers were trained. The pretrained layers consisted of Dense layers with 1024, 512, 256 and 128 perceptrons, with each layer being followed by an appropriate Batch Normalization layer. All the perceptron layers are activated using a ReLu activation function and the optimization technique used while training is Adam Optimization, much like the previous model.  The initial model was trained for 20 epochs with a batch size of 128, while on the other hand when the model was used for transfer learning and trained on Pavia dataset only 10 epochs with 128 batch size were performed. 
\section{Results}
\subsection{CNN Model}
\begin{figure}[ht]
  \centerline{\begin{subfigure}{0.3\textwidth}
    \centering
    \frame{\includegraphics[width=.6\linewidth]{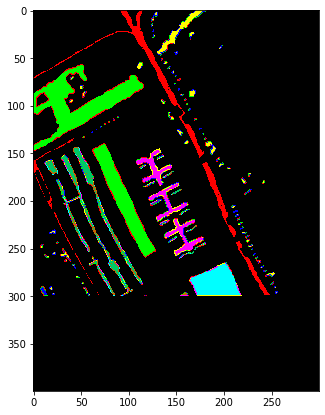}}
    \caption{Ground Truth}
    \label{fig:gtPavia}
  \end{subfigure}
  \begin{subfigure}{0.3\textwidth}
    \centering
    \frame{\includegraphics[width=.6\linewidth]{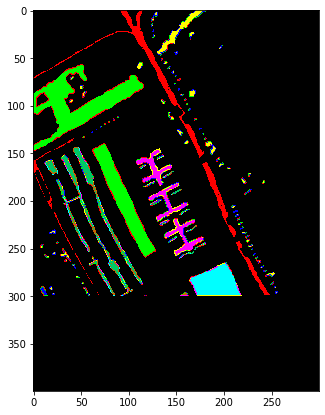}}
    \caption{Prediction}
    \label{fig:predictedPavia}
  \end{subfigure}}
  \caption{University of Pavia dataset}
  \label{fig:PaviaUniversity}
\end{figure}

Firstly, the Indian Pines dataset is applied to the CNN model.Test set accuracy came out to be \textbf{99.31}\% which is quite high, but the model had to be trained for 100 epochs which took around 15 minutes to run on Google Colaboratory. This trained model was saved to be used for transfer learning purposes in future.
\begin{figure}[h]
    \centering
    \frame{\includegraphics[scale=0.35]{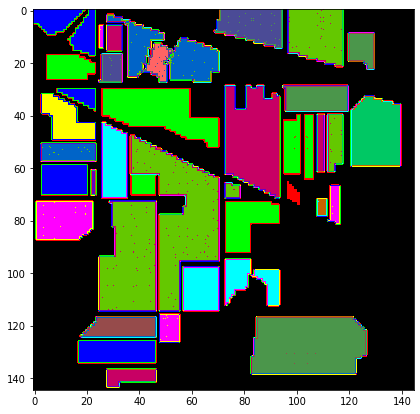}
  }
  \caption{Indian Pines dataset}
  \label{fig:IndianPines}
\end{figure}
For training over the Pavia University dataset, the above model was loaded and the last few layers were popped off, as they were specifically trained on Indian Pines Dataset. New layers were added so that they could be trained specifically on the Pavia University dataset.The test set accuracy for the Pavia University dataset came out to be \textbf{99.93}\% and it had to be trained only for 3 epochs that took approximately 30 seconds, which is significantly better than the last model. The prediction made are shown in Fig \ref{fig:predictedPavia}.

\subsection{MLP Model 1}
The MLP model was a very basic model with varying activation functions, and it was first trained on the Indian Pines dataset.\\

It can be seen that the average accuracy of the model is 99.70\% which is acceptable, and it should be noted that since the model was not having too many hidden layers it only took around 10 minutes to be trained for 20 epochs on Google Colaboratory even at high configurations.\\
Once we had the above trained model, we used the same for transfer learning by editing the last layer, and adding new trainable layers with trainable parameters. The model hence was then trained on the Pavia dataset, upon whose training and validation.

It should be noted that during the transfer learning the model was trained only for 10 epochs and the accuracy that we received out of the model on the Pavia dataset is close to 96.78\% which is acceptable when we see in the context of transfer learning. It should be noted that the accuracy decreased before and after applying the transfer learning because the initial layer was still trained with weights of the previous Indian Pines dataset, and this decrease in accuracy on applying transfer learning algorithm is very commonly observed.

\subsection{MLP Model 2}
The MLP model was first trained on the Indian Pines dataset.
\par In the  above figure, we can see that the average accuracy of the model is 99.95\% which is quite high, but it should be noted that the model had to be trained for 20 epochs which took around 10 minutes to run on Google Colaboratory even at such high configurations.This trained model was saved to be used for transfer learning purposes in future.\\
Once we had the trained model, we used the weights of that model for the transfer learning purpose as it would not only reduce the time it takes for the model to train on the Pavia dataset, which was substantially larger than the Indian Pines dataset.
\par It should be noted that during the transfer learning the model was trained only for 10 epochs and the accuracy that we received out of the model on the Pavia dataset is close to 98.71\% which is still quite high when we see in the context of transfer learning. The accuracy was also a substantial leap from that of the basic model, and hence we can say that this model out-performed the previous model.

\subsection{MLP Model 3}

This MLP model was similar to a horizontally scaled architectured form of the second model, and it was also first trained on the Indian Pines dataset.
\par In the  above figure, we can see that the average accuracy of the model is 99.90\% which is still quite high, but it should be noted that the last model had already achieved a greater accuracy, and this model in turn had also to be trained for 20 epochs which took a greater time, about 15 minutes to run on Google Colaboratory even at such high configurations.This trained model was again saved to be used for transfer learning purposes in future.\\
Once we had the trained model, much like we did with the previous two models, we used the weights of that model for the transfer learning purpose on the Pavia dataset.

It should be noted that during the transfer learning the model was trained only for 10 epochs and the accuracy that we received out of the model on the Pavia dataset is close to 98.19\% which is quite lower than what we had already achieved. This was probably seen because the pre-trained model had three large perceptron layers that were not trained during transfer learning, so the model was actually overfitting the Indian Pines dataset and the extra few layers that were later trained on the Pavia dataset, were not able to align the weights and capture the true features.
\vspace{0.3cm}\\
\textbf{Comparison}\\
Now since we have seen the separate outputs of the three different MLP models, we also need to compare the three to decide the one whose architecture best fits our objective. In the table \ref{tab:mlpTable}, we have compared the three models in a tabular form, and though by looking at the output predicted images, being it of the Indian Pines or the Pavia dataset, all three seems to be exactly the same, but when we see the test accuracy and the average accuracy store we find a difference there. The first model being a very basic model is out-performed by the second model, which though requires a higher computational power and time for training but also provides a higher accuracy. In case of the third model, we saw that mere horizontally scaling and increasing the number of perceptron layers did not result in increase in the accuracy but rather decreased the accuracy due to overfitting over the dataset and the same can be seen when the model was tested after transfer learning.
\begin{table*} 
\centering
\resizebox{\textwidth}{!}{\begin{tabular}{|c|c|c|c|}
\hline
\bfseries Model & MLP Model 1 & MLP Model 2 & MLP Model 3 \\
\hline
\bfseries Layers & 3 & 4 & 7 \\
\hline
\bfseries Architecture & (18750,10000,5000,9) & (18750,472,168,72,9) & (18750,1024,512,256,128,72,32,9) \\
\hline
\bfseries Trainable Params & 50,050,009 & 12,825 & 11,921 \\
\hline
\bfseries IP Test Accu. & 99.70\% & 99.95\% & 99.90\% \\
\hline
\bfseries IP Avg. Accu. & 99.59\% & 99.96\% & 99.78\% \\
\hline
\bfseries Pavia Test Accu. & 96.78\% & 98.71\% & 98.19\% \\
\hline
\bfseries Pavia Avg. Accu. & 95.37\% & 97.84\% & 96.83\% \\
\hline
\end{tabular}}
\caption{Comparison of MLP models}
\label{tab:mlpTable}
\end{table*}

\begin{figure*}[ht]
  \centerline{\begin{subfigure}{0.3\textwidth}
    \centering
    \frame{\includegraphics[width=.7\linewidth]{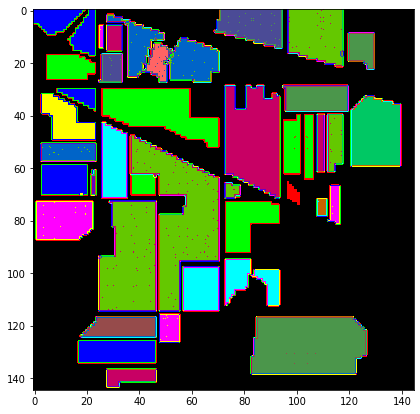}}
    \caption{MLP model 1}
    \label{fig:}
  \end{subfigure}
  \begin{subfigure}{0.3\textwidth}
    \centering
    \frame{\includegraphics[width=.7\linewidth]{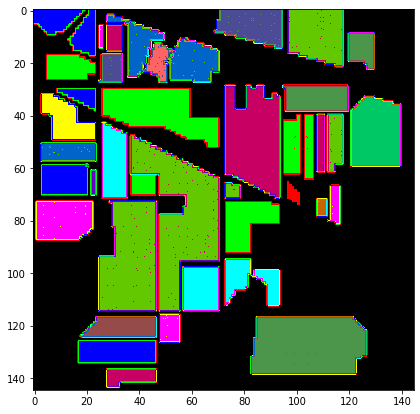}}
    \caption{MLP model 2}
    \label{fig:}
  \end{subfigure}
  \begin{subfigure}{0.3\textwidth}
    \centering
    \frame{\includegraphics[width=.7\linewidth]{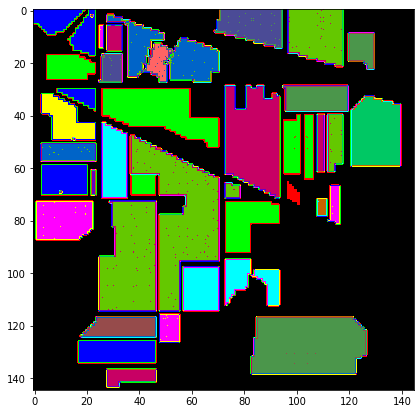}}
    \caption{MLP model 3}
    \label{fig:}
  \end{subfigure}}
  \caption{Predicted Images of Indian Pines dataset}
  \label{fig:IPOutput}
\end{figure*}
\begin{figure*}[ht]
  \centerline{\begin{subfigure}{0.3\textwidth}
    \centering
    \frame{\includegraphics[width=.7\linewidth]{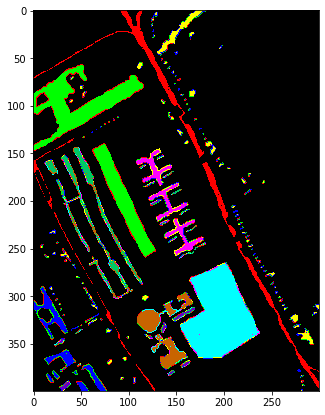}}
    \caption{MLP model 1}
    \label{fig:}
  \end{subfigure}
  \begin{subfigure}{0.3\textwidth}
    \centering
    \frame{\includegraphics[width=.7\linewidth]{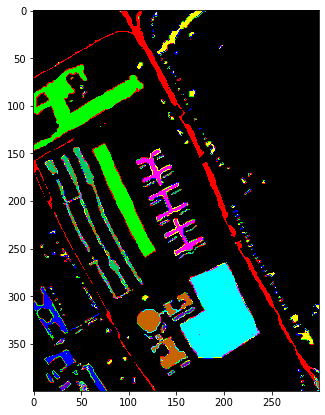}}
    \caption{MLP model 2}
    \label{fig:}
  \end{subfigure}
  \begin{subfigure}{0.3\textwidth}
    \centering
    \frame{\includegraphics[width=.7\linewidth]{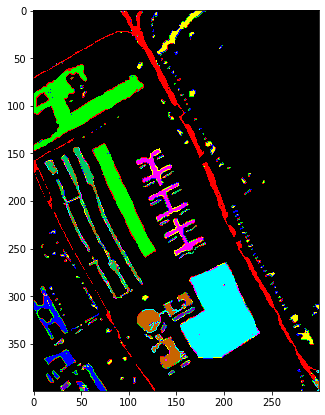}}
    \caption{MLP model 3}
    \label{fig:}
  \end{subfigure}}
  \caption{Predicted Images of Pavia University dataset}
  \label{fig:PVOutput}
\end{figure*}
With the above comparison, it is conclusive that the second MLP model that we proposed is the better of the three, with a test accuracy of 99.95\% on Indian Pines and of 98.71\% on the Pavia Dataset. When we compare the above result with the one that we had obtained while addressing the same problem by using a CNN based transfer learning approach, where we obtained a test accuracy of  99.91\% on Indian Pines and 99.93\% on the Pavia dataset, we can conclude that the MLP model though was faster in terms of training and the convergence to the optimum result was faster than what it took in case of CNN. The CNN also captures the spatial features of the image dataset, apart from just the spectral ones, whereas the MLP model only captures the spectral features. 
\section{Conclusion}
The MLP based models are straight-forward and are used in a wide range of image classification objectives. Unlike the CNN, they require a smaller training time since there are no tasks to capture the spatial features like convolution or pooling. Scaling the MLP architecture, either vertically or horizontally does lead to increase in the accuracy of the model, at the cost of time, but there is an upper limit to it, after which the model starts to experience overfitting and performs worse in the case of transfer learning. These facts were evidently seen with the detailed comparison of the three different MLP models that were implemented and discussed.
\par The use of transfer learning substantially decreases the training time required for the model, on larger datasets like Pavia by extracting weights and knowledge from smaller datasets like the Indian Pines dataset. The crux of transfer learning lies in fact that we take advantage of the inner layers trained in the previous model and use it’s general information (in our case about the hyperspectral images) to train the outermost few layers. Thus, saving time by not having to train the inner layers again.\\
One more argument that arose was whether a CNN based transfer learning model or a MLP based transfer learning model.The decision of which model to prefer is based on the application and objective and also may vary with the available resources. In case when the terrain classification is to be used for purposes like agricultural research, the spatial features need to be taken care of and hence a CNN based model would work better in this case. On the other hand, in applications involving military remote sensing the spectral features overlook the spatial ones, and hence the MLP based model should be preferred over CNN in this case.



\section*{References}
\printbibliography[heading=none]

\end{document}